\relax
\documentclass[letterpaper]{article}
\usepackage{named}
\usepackage[colorlinks=true,linkcolor=black,anchorcolor=black,citecolor=black,filecolor=black,menucolor=black,runcolor=black,urlcolor=black]{hyperref}
\usepackage{url}
\usepackage{graphicx}
\usepackage{amssymb}
\usepackage{tabularx,booktabs}
\newcolumntype{Y}{>{\centering\arraybackslash}X}
\newcolumntype{Z}{>{\raggedleft\arraybackslash}X}

\usepackage{xspace}
\usepackage{color}
\usepackage[spanish,english]{babel}

\newcommand{\no}[1]{}
\usepackage[usenames,dvipsnames]{xcolor}
\newcommand{\todo}[1]{}

\title{Goal Recognition via Model-based and Model-free Techniques}
\author{Daniel Borrajo\thanks{J.P.Morgan AI Research, New York, NY
    (USA). On leave from Universidad Carlos III de Madrid. The
    position at the lab is as a
    consultant. daniel.borrajo@jpmchase.com},
  \hspace*{0.2cm} Sriram Gopalakrishnan\thanks{Arizona State University. sgopal28@asu.edu} \and
  Vamsi K. Potluru\thanks{J.P.Morgan AI Research, New York, NY (USA). vamsi.k.potluru@jpmchase.com}
  }

\date{}

\begin{document}

\maketitle

\begin{abstract}
    Goal recognition aims at predicting human intentions from a trace of observations. This ability allows people or organizations to anticipate future actions and intervene in a positive (collaborative) or negative (adversarial) way. Goal recognition has been successfully used in many domains, but it has been seldom been used by financial institutions. We claim the techniques are ripe for its wide use in finance-related tasks. The main two approaches to perform goal recognition are model-based (planning-based) and model-free (learning-based). In this paper, we adapt state-of-the-art learning techniques to goal recognition, and compare model-based and model-free approaches in different domains. We analyze the experimental data to understand the trade-offs of using both types of methods. The experiments show that planning-based approaches are ready for some goal-recognition finance tasks.
\end{abstract}

\section{Introduction}

Humans interact with the world based on their inner motivations (goals) by performing actions. Those actions might be observable by financial institutions. In turn, financial institutions might log all these observed actions for better understanding human behavior. Examples of such interactions are investment operations (buying or selling options), account-related activities (creating accounts, making transactions, withdrawing money), digital interactions (utilizing the bank's web or mobile app for configuring alerts, or applying for a new credit card), or even illicit operations (such as fraud or money laundering). Once human behavior can be better understood, financial institutions can improve their processes allowing them to deepen the relationship with clients, offering targeted services (marketing), handling complaints-related interactions (operations), or performing fraud or money laundering investigations (compliance)~\cite{icaif20}.

The field of goal recognition lies at the crux of understanding human behavior~\cite{Carberry01,KautzAllen86,Sukthankar14}. Given a trace of observations of some agent taking actions in the environment, goal recognition techniques try to infer the agents' goals. Similarly, plan recognition will try to infer the next actions (plan) that the agent will execute~\cite{KautzAllen86}, and activity recognition will infer the current activity of the agent~\cite{jdsn13}. These three tasks are highly related, though the objective of the recognition problem is different; agents come up with goals they would like to achieve for which they generate plans composed of sequences of actions~\cite{planning-book}. However, these tasks are not exactly equivalent. For instance, there might be more than one plan to achieve a goal and more than one goal achievable by a plan, so even if one recognizes what plan an agent is pursuing, that might not uniquely identify the goals the agent is trying to achieve, and vice versa. In this paper we focus on goal recognition.

Goal recognition by an agent aims at inferring the goals of another agent from observation. A convenient way to model this process has been in terms of using Bayesian models that compute a posterior probability of some goals/plans based on prior probabilities and new observations~\cite{bui2002policy,PynadathWellman95,ramirez10}. Other alternatives are based on creating plan libraries and matching new observations with those libraries~\cite{KautzAllen86,avrahami-zilberbrand05fast}. The latter kind of techniques require careful curation of those libraries, which can result in a time-consuming effort and is limited by the set of libraries used. Some works have automated the construction of those libraries~\cite{Mooney90}. However, often the search in the space of plans can result in a huge computational effort due to the exponential number of potential plans when the size of the action space increases~\cite{kautz-thesis}.

More recently, automated planning techniques have been used to infer an agent's goals~\cite{holler2018plan,ramirez10,PereiraOM20,yolanda2015fast,sohrabi2016plan,vered2017heuristic}. They replace the plan libraries by using a domain model. The techniques based on automated planning provide good performance, are domain-independent and are provably sound. However, they require again to manually define the underlying planning model (domain and problem descriptions) and rely on the correctness of the model. The modeling effort of these approaches varies from STRIPS planning models~\cite{ramirez10,PereiraOM20,yolanda2015fast} to more knowledge-intensive as Hierarchical Task Networks (HTNs)~\cite{holler2018plan}. 

Another approach to perform goal recognition consists of using state of the art machine learning approaches to learn patterns of actions that predict goals. So, instead of requiring an action model, they require training instances.

These two approaches to goal recognition are based on the two major approaches to AI~\cite{geffner2018model}: model-based (e.g. planning, search) and model-free (learning). A key distinction between the two approaches in relation to goal recognition relates to the assumption that planning approaches make about agents' rationality: they will try to achieve their goals in the best possible way (optimal). On the contrary. model-free approaches do not have to assume that the agents' behavior is rational. Some authors have previously shown lack of rationality and noisy decisions in specific domains, as games~\cite{min2016}. 

In this paper, we provide a comparison of state of the art domain-independent model-based approaches~\cite{PereiraOM20} and the adaptation to goal recognition of two state of the art model-free approaches: long short-term Memory (LSTM)~\cite{hochreiter1997long} and XGBoost~\cite{chen2016xgboost}. Previous work on model-based goal recognition showed impressive results in terms of accuracy of recognized goals, close to 100\%, with very few observations. We claim this is due to the use of simple benchmark cases. Therefore, we have created a few harder goal recognition tasks. In particular, we have created a benchmark focusing on a finance-related application to show its potential impact on this area. In this new domain, bank clients can open accounts, receive their payrolls and make payments to buy items or pay for their kids college. Financial institutions can partially observe all these actions (traces of human behavior), as described in~\cite{icaif20}. In this domain, the task of a goal recognition system is to determine for which products or services humans will be making payments by analyzing previous observations.



In the next sections, we will present related work, introduce the learning-based approaches, show and discuss the experiments and draw some conclusions.

\section{Related Work}

We have covered in the introduction the closest model-based works related to goal recognition~\cite{holler2018plan,ramirez10,PereiraOM20,yolanda2015fast,sohrabi2016plan,vered2017heuristic}. In the case of model-free (learning) approaches, some authors have started recently using machine learning techniques based on Convolutional Neural Networks (CNNs) combined with plan libraries~\cite{granada2020haprec} or LSTM~\cite{amado2019latrec} to infer goals from a sequence of observations. In these works, observations are images that represent states when some task is being solved. Our observations are not required to take the format of an image and they are actions performed by the agent instead of states; states contain more information than just the actions. We argue that action traces (sequences) match more closely financial transaction traces, and they are thus a better fit for goal recognition in finance~\cite{icaif20,workshop-icaps20}.

There has also been increasing interest in several domains to devise goal recognition systems, such as the work on story understanding~\cite{Charniak1993}, network security~\cite{Geib2009}, or computer games~\cite{hooshyar2018data}. The latter domain can be considered similar to some financial tasks, the adversarial ones, such as fraud, money laundering or market based. In games, some examples used a combination of model-based (using manually generated Markov Logic Networks) with model-free approaches that learned the associated probabilities~\cite{ha2011}. Other authors used LSTMs to predict user's goals in game playing~\cite{min2016}. The input to the LSTM includes the previously achieved subgoals, as well as the executed action. They also assumed agents do not interleave working on several goals, since the goals they handle do not share anything in common. However, in many situations (e.g. all domains considered in planning-based approaches), goals are sets of propositions and goals can have some shared propositions. Additionally plans/action traces can approach multiple goals before settling on one or the other. So the assumption of non-interleaving plans is very constraining.

As an example, suppose that there are three propositions in a financial application :

\begin{center}
$P1$=\texttt{(paid Client Car)},\\
$P2$=\texttt{(paid Client House)} and\\
$P3$=\texttt{(has-credit-card Client)}.
\end{center}

Then, one goal could be G1=$\{P1,P3\}$ and another one could be G2=$\{P1,P2\}$. While both goals are different, they do
share some common parts. Therefore, the assumption that agents cannot interleave work on them is too restrictive, since
the client could be trying to reach both by pursuing $P1$. Instead, we allow goals to share common components, so agents
can potentially be working on two goals at the same time, until they commit finally to one of them.

Also, as work in other goal recognition domains, their work is domain dependent. Adapting it to a new domain would require some feature modeling effort as they mention in the paper. Our use of machine learning techniques is domain-independent and thus is a better comparison to model-based methods. Our LSTM approach returns the predicted goal after taking as input a sequence of one-hot encoding of actions (which are the observations). This approach makes it independent of the domain. 

\section{Background and Problem Definition}

We define the goal recognition framework as a tuple $GR=\langle F, A, \mathcal G\rangle$, where $F=\{f_1,\ldots,f_n\}$ is a
set of propositions, $A=\{a_1,\ldots,a_m\}$ is a set of instantiated actions, and ${\mathcal G}=\{G_1,\ldots, G_p\}$ is a set of
goals. In automated planning, a proposition is an instantiated literal. For instance, \texttt{account-owner-C1-A1} would represent the fact that a client \texttt{C1} is the owner an account \texttt{A1} in a bank in a financial domain. Similarly, an instantiated action is traditionally composed of an action name and constants as parameters. For instance, \texttt{open-account-C1-A1} would be an instantiated action where the client \texttt{C1} opens the account \texttt{A1} in the same domain. Each goal $G_i\in \mathcal G$ is a set of propositions in $F$; that is,
$\forall g_j\in G_i, g_j\in F$. In the example domain, a potential goal $G_i$ could be: $G_i=\{$\texttt{paid-C1-House1,paid-C1-College1}$\}$ representing the goals of buying a house and paying for the children college.

The learning algorithms to be defined in the next section receive as input a set of training instances and return a goal-recognition classifier. Each training instance is a tuple $i=\langle O, G\rangle$. $O=\{o_1,\ldots,o_o\}$ is a sequence of observations of instantiated actions ($\forall o_i, o_i\in A$), and $G\in \mathcal G$. A goal-recognition classifier is an algorithm that takes as input an observation sequence $O'$ and returns a label that is a goal $G$ in $\mathcal G$. Note that the learning technique does not take as input a domain model (actions and predicates), nor any prior information, as other planning-based goal-recognition approaches~\cite{ramirez10}. So, each action is just a label without any semantics. In fact, in order to properly work with the proposed learning techniques, we used numerical values for those labels, as well as for the goals. 

As an example of an observation sequence in the hypothetical financial domain, we could have observed: 

\begin{tabbing}
123\=\kill
\>$\langle$\= \texttt{create-account-C1-A1,work-C1,\ldots,work-C1,}\\
\>\> \texttt{pay-C1-House1}$\rangle$ 
\end{tabbing}

The action \texttt{work} increases the balance of the account. And the \texttt{pay} action achieves the goal (paying House1). The training instance could be something like:

\[\langle \langle 11,23,23,\ldots,23,\rangle, \{35\}\rangle\]

where 11 represents \texttt{account-owner-C1-A1} and similarly for the other literals, while 35 represents \texttt{pay-C1-House1}. These integer values are replaced by a one-hot encoding.

\section{Learning Models for Goal Recognition}

We apply two state of the art model-free methods for learning goals, namely recurrent neural networks (RNNs) and gradient boosted trees (GBTs). RNNs have been applied to a wide-range of problems including neural machine translation~\cite{cho2014learning} and
sequence to sequence learning~\cite{sutskever2014sequence}. They apply the same set of weights for each action of the sequence which makes them memory efficient. It also provides them translational symmetry. 
LSTMs are a type of RNN with memory and gating that can handle longer sequences of data~\cite{hochreiter1997long}. A recent version has been widely adopted to handle even longer input sequences~\cite{cho2014learning}.

Gradient boosting is used to learn an ensemble model combining weak learners to create a good classifier. The weak learners
are typically decision trees and hence the term GBT. At a high-level, gradient boosting fits a new tree at each step to
the residual of the current pseudo-residuals. A popular implementation of GBT with additional tuning, such as proportional
shrinking of leaves and Newton boosting, is provided by XGBoost~\cite{chen2016xgboost}.

\section{Experiments}

In this section, we present the experimental setting, the results obtained and an in-depth analysis.

\subsection{Settings}
Our experiments were conducted on domains used in previous works on goal recognition, such as three International Planning Competition (IPC\footnote{\url{ipc.icaps-conference.org}}) domains: Block-words, Logistics, and a Simple Grid domain. We also ran experiments on a new domain inspired on a finance task, which we simply call Buy-domain. We will now describe each of the domains in more detail.

\begin{figure}
    \centering
    \includegraphics[scale=0.3]{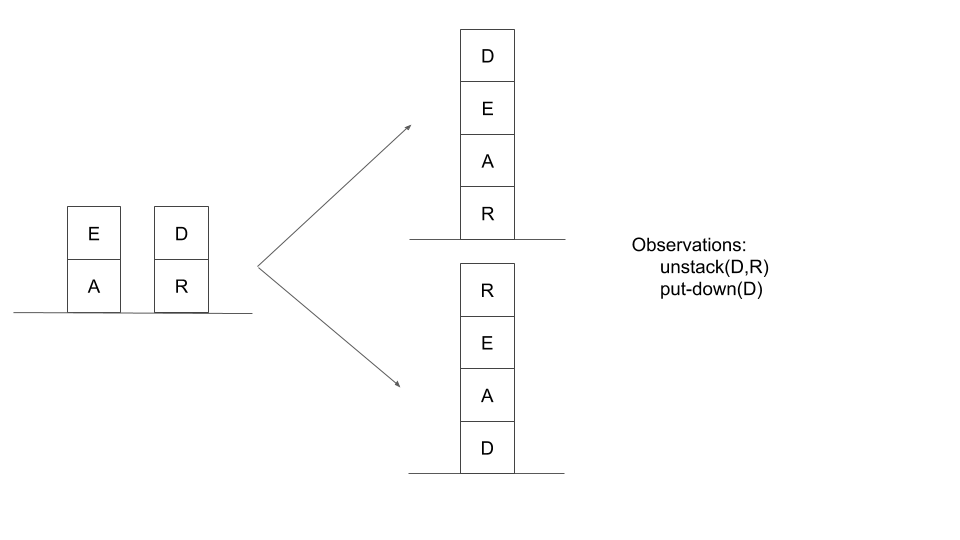}
    \caption{Example of an initial state in the Blocks-words domain (left) and two potential goals (middle). It also shows some observations (actions) on the right.}
    \label{fig:bw}
\end{figure}

\textbf{Block-words} is equivalent to the known Blocksworld domain, but involves blocks whose names correspond to letters. Blocks start in a specific configuration on a table. A robotic arm can unstack/stack blocks from/to others or pick-up/put-down blocks from/on the table, in order to reach a specified goal configuration that represents a specific English word. As an example, given the blocks named R, A, E, D, we could create problems whose goals are to create towers of blocks with the words READ, DEAR, or RED. Figure~\ref{fig:bw} shows an example of an initial state (left) and two possible goals (right): READ and DEAR. It also shows a partial observation (two actions). The goal would be to predict which of the two words the agent is trying to create given those observations.


We generated two sets of instances in this domain. In the first one, there are two goals composed of 24 propositions each (blocks forming a big tower). The only difference between the two goals is the block on top. The initial state was randomly generated. This is an example of a type of task that is hard for goal recognition, since it is not until the end that the observer can really decide which goal the other agent is pursuing (which block will be on top). Also, in order to test the ability of learning some prior probability distributions, the probability of selecting one of the words was 80\% and the probability of selecting the other one 20\%. We would expect that techniques that can handle this prior probabilities or learn them from data would have a higher accuracy from the start of the observation sequence. The second set of instances was composed of five goals (words) and for each generated problem we picked randomly which one should be the problem's goal. The initial state was also randomly generated. This task was designed to test the generalization capability of the goal recognition systems (especially the learning-based ones). In order to train the learning-based systems, we randomly generated 10,000 problems that were solved by a sub-optimal planner (the first iteration of the {\sc lama} planner~\cite{lama}). The generated plans were used as training traces together with their corresponding goals.

\begin{figure}[hbt]
    \centering
    \includegraphics[scale=0.3]{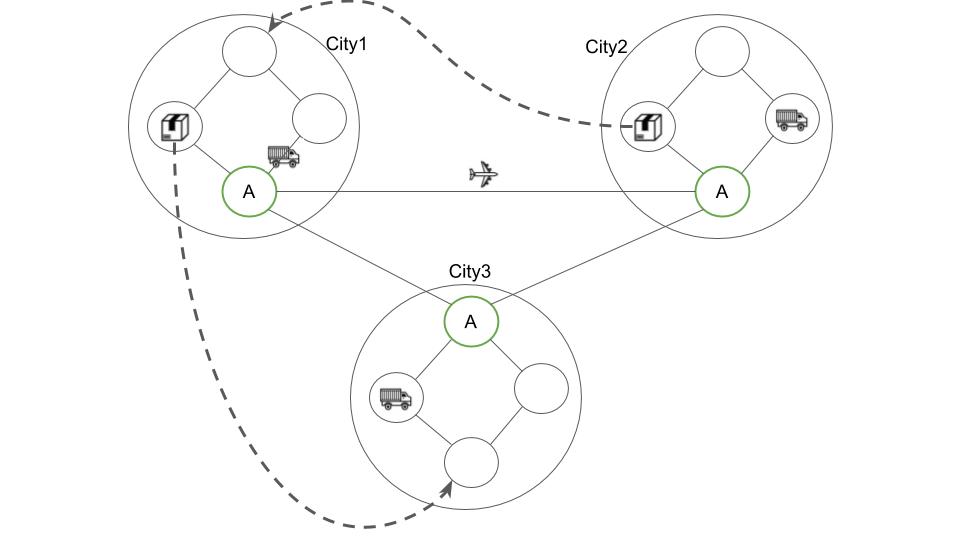}
    \caption{Example of an initial state in the logistics domain that has 3 cities, with 4 locations each.}
    \label{fig:logistics}
\end{figure}

\textbf{Logistics} involves transporting a set of packages from their initial locations to specified goal locations for each package. Packages can be transported within a city via a truck, and across cities via airplanes; this is illustrated in Figure \ref{fig:logistics}. In our experiments there were 10 cities, with 4 locations within each city. Each city had 1 truck, and there was 1 airplane to move packages within cities. There were 10 packages in total, which had to be moved from their initial location to their respective goal locations. We used two settings in this domain. One setting involved two goals and a fixed initial state; the likelihood of one goal was $80\%$ and the other was $20\%$. Since the plans started with the packages in the same initial location, the prefixes of the plans reaching the goal state had many shared actions. 
The other setting of the logistics domain involved 10 goals, and random initial states. For the first setting, we randomly generated 2,000 problems and for the second setting we randomly generated 10,000 problems. Again, these problems were solved by the same planner as before to generate the training traces.

\begin{figure}[hbt]
    \centering
    \includegraphics[scale=0.3]{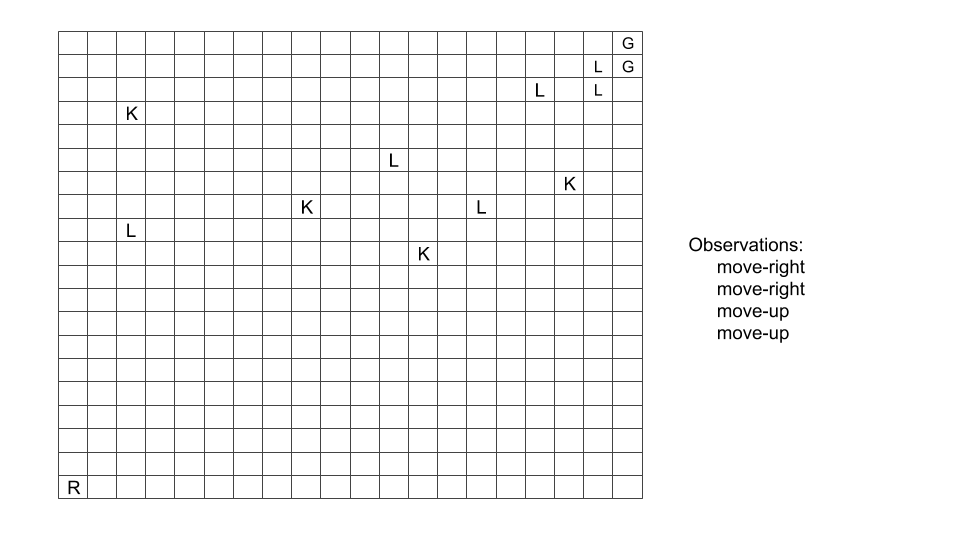}
    \caption{Example of an initial state in the Grid domain. R represents the initial state (position of the robot), K are keys, L are locked positions, and G are potential goals. It also shows some observations (actions) on the right.}
    \label{fig:grid}
\end{figure}

\textbf{Grid} involves a robot that can move in a grid. Some tiles are locked and the robot has to use keys that are randomly distributed in the grid to open those tiles. The goal is to move the robot from its initial position to a final tile. Figure~\ref{fig:grid} shows an example of an initial state of the robot (bottom left marked with R), the grid, a set of keys (K) and locked positions (L). It also shows two possible goals (upper right) marked with G. At the right of the figure, there is a sequence of four observations (actions).

In this domain we generated two sets of examples. In the first one, we selected the initial state and goals as in Figure~\ref{fig:grid} and some locked positions and keys randomly distributed. This task was designed again with the objective of making the goal recognition task difficult since the prefix of plans (first actions of the plan) will be similar to plans achieving the two goals. Again, we biased the goals with a 80/20\% probability distribution to test the ability to learn the priors. In the second test, we generated random initial states and goals (from a set of 20 pre-defined goals). We also used 10,000 traces for each setting to train the learning approaches.

\textbf{Buy} domain is a simplified financial domain. Clients of a bank can open different kinds of accounts, work, receive their payrolls, move funds from one account to another and buy products (e.g. house) or pay for their kids education. In the initial state, the clients open an account with zero balance and when they have worked for some time and they have enough money in the balance, they can either buy products or pay for the college. This domain cannot be handled by model-based approaches easily since it includes numeric variables that are not handled by most goal recognition systems currently. Figure~\ref{fig:buy} shows an observation trace of this domain.

\begin{figure}[hbt]
    \centering
    \begin{tabular}{l}
        create-account Client1 Account1 \\ 
        payroll Client1 \\ 
        payroll Client1 \\ 
        \ldots \\ 
        payroll Client1 \\ 
        buy Client1 Product1 \\ 
        payroll Client1 \\ 
        \ldots \\ 
        payroll Client1 \\ 
        payroll Client1 \\ 
        pay-college Client1 \\ 
    \end{tabular}
    \caption{Example of an observation trace in the Buy domain.}
    \label{fig:buy}
\end{figure}

In this domain we also generated two sets of tasks. In the first task, we generated two potential goals (either buy a house or pay for the kids college) with a prior probability of 80/20\%. In those problems where the goal was to buy a house, we added an artificial first action to the observations of opening a savings account. This action does not achieve any of the goals, so the planning-based approaches cannot guess the connection to the goals. This kind of task shows that when there is a relation between arbitrary irrelevant actions, planning-based approaches will not be able to correctly detect the right goal from the start. These actions are common to all real world domains where agents are humans, given that often they do not behave rationally. Instead, learning approaches will be able to detect those correlations and boost the accuracy in those tasks. The second task has 10 goals (buying different products or attending different colleges) and random initial states.

\subsection{Training and Test Instances}
The training data was generated using the first iteration of the LAMA sub-optimal planner~\cite{richter2010lama}. For the techniques that involve learning, we used a $80/20$ split of the data (traces) into training and validation. In order to compare with the landmark goal recognition technique, we kept a dedicated set of 100 test traces for each setting separately (not in the $80/20$ data set). Table~\ref{tab:summary} shows a summary of some key metrics with respect to the training instances.

In our experiments, we tried to intentionally generate problems that are hard for goal recognition. We did this by having some goals close to each other, so plans would share a large prefix, and shared actions would make goal recognition difficult. Additionally, having some settings with random and diverse initial conditions would make it hard to learn patterns of actions (subsets or sub-sequences of actions) that correlate to goals.

\begin{table}[hbt]
    \centering
    \begin{tabular}{c|c|c|c|c}
         Domain & Set & Max  & Num. & Num.  \\ 
         &  &  length &  actions &  goals \\ \hline
         Blocks-words& Set1 & 434 & 424 & 2\\ 
         Blocks-words& Set2  & 16 & 24 & 5\\ 
         Logistics& Set1 & 67& 79 & 2\\ 
         Logistics & Set2 & 111 & 846 & 10\\ 
         Grid& Set1 & 424& 424& 2\\ 
         Grid& Set2 & 33& 1212& 10\\ 
         Buy& Set1 & 43 & 6 & 2\\ 
         Buy& Set2 & 541 & 25 & 10\\ 
    \end{tabular}
    \caption{Summary of key characteristics of the training instances.}
    \label{tab:summary}
\end{table}

\subsection{Model-Based and Learning Methods}
To evaluate the performance of goal recognition with LSTMs and XGBoost, we compared the accuracy with two other baseline techniques based on Landmark-based goal recognition using planning~\cite{PereiraOM20} (which we will call \textit{LGR}). In the first version, we used their code. In that version, they do not contemplate prior probabilities of some goals, as early work on goal recognition by planning did~\cite{ramirez10}. So, in the second version we modified their work to factor in the prior probability distribution of goals when making the prediction. We selected to compare against LGR given that their results are better or similar to the rest of current state of the art algorithms based on planning. We used the code made publicly available by the authors.\footnote{\url{https://github.com/ramonpereira/Landmark-Based-GoalRecognition}}
To compare with the baseline \textit{LGR} method, we used the same method reported by Pereira et al.: reporting the accuracy results after some percentage of observations were made. We used the same proportions as in their work, viz $\{0.1,0.3,0.5,0.7\}$. We set the threshold parameter to $0.1$ for their \textit{LGR} method.

For LSTMs, we set the embedding dimension of actions to $512$ to cover the cases where we had a large number of actions.
The learning rate was $0.01$ and both batch sizes for training and hidden dimension were set to $32$. The number of
epochs for training was $10$.
These parameters were not tuned to the datasets. We would expect better performance if tuned to the problem. 
For GBT, we learned $100$ trees with a maximum depth of $3$.

\section{Discussion}
Table~\ref{tab:results} shows the results in terms of accuracy and time to make predictions for the 100 test plan traces. In all domains except Logistics, learning-based methods outperform the model-based LGR method. Factoring the prior into the prediction of LGR only made sense when there were two goals, because those were the settings when the goal distribution was skewed (80/20). Otherwise, they were uniform and the priors should not make a difference. Factoring the prior into LGR only seemed to hurt the prediction since the LGR prediction mechanism had nothing to do with distribution and learning; rather LGR used the landmarks observed to predict. The only case when the prior distribution helped the accuracy was in the Grid domain, where LGR was performing abysmally (there are no landmarks or they are very uninformative in that domain). We now analyze the results domain-by-domain.



\begin{table*}[hbtp]
  \hspace*{-2cm}
  \begin{minipage}{1.2\textwidth}
  \small
  \begin{tabularx}{1.2\textwidth}{cc *{8}{Z}}
        \toprule
         \multicolumn{1}{c}{\ } & \multicolumn{1}{c}{\ } & \multicolumn{2}{c}{LGR}&
         \multicolumn{2}{c}{LGR w Prior}&\multicolumn{2}{c}{XGBoost}&  \multicolumn{2}{c}{LSTM}
         \\
         \multicolumn{1}{c}{Domain} & \multicolumn{1}{c}{Observ.} & \multicolumn{1}{c}{Acc.} & \multicolumn{1}{c}{Time} & \multicolumn{1}{c}{Acc.} & \multicolumn{1}{c}{Time} & \multicolumn{1}{c}{Acc.} & \multicolumn{1}{c}{Time} & \multicolumn{1}{c}{Acc.} & \multicolumn{1}{c}{Time}\\
     \midrule
        Blockswords & 10 & 69.0 & 160.81 & 77.0 & 126.87 & 79 & 0.03 & 79 & 0.02\\
         2 goals & 30 & 76.0 & 147.75 & 77.0 & 137.23 & 92 & 0.03& 79& 0.02 \\ 
          & 50 & 86.0 & 155.22 & 77.0 & 151.40 &97& 0.03& 79& 0.02\\
          & 70 & 86.0 & 147.35 & 77.0 & 140.75 & 96& 0.03& 79& 0.02\\    
        Blockswords & 10 & 23.0 & 28.96 & 23.0 & 28.96 &41 & 0.001& 12& 0.03\\
         5 goals & 30 & 23.0 & 29.32 & 23.0 & 29.32 & 41& 0.001& 24& 0.03\\    
           & 50 & 44.0 & 29.17 & 44.0 & 29.17 & 64& 0.001& 31& 0.03\\    
          & 70 & 64.0 & 29.93 & 64.0 & 29.93 & 64& 0.001 & 78& 0.03\\    \midrule 
         Logistics & 10 & 68.0 & 417.89 & 81.0 & 426.75 & 75.0 & 0.001 & 79.2 & 0.06\\
         2 goals & 30 & 81.0 & 461.64 & 81.0 & 415.65 & 75.0 & 0.001 & 79.2 & 0.06\\
                 & 50 & 81.0 & 398.90 & 81.0 & 462.80 & 75.0 & 0.001 & 79.2 & 0.06\\
                 & 70 & 81.0 & 405.61 & 81.0 & 429.85 & 87.0 & 0.001 & 79.2 & 0.06\\     
         Logistics & 10 & 10.0 & 1657.51 & 10.0 & 1657.51 & 14.0 & 0.001 & 11.79 & 0.02\\ 
         10 goals & 30 & 24.0 & 1668.00 & 24.0 & 1668.00 & 44.0  & 0.001 & 21.30 & 0.02\\    
                  & 50 & 50.0 & 1283.90 & 50.0 & 1283.90 & 45.0 & 0.001 &  47.78 & 0.02\\    
                  & 70 & 69.0 & 1300.52 & 69.0 & 1300.52 & 47.0 & 0.001 &  49.12 & 0.02\\     \midrule
         Grid & 10 & 1.0 & 115.07 & 77.0 & 78.99 & 77& 0.001& 78& 0.03\\
         2 goals & 30 & 1.0 & 96.01 & 77.0 & 80.35 & 94& 0.001&94 & 0.03\\    
                & 50 & 1.0 & 104.15 & 77.0 & 94.98 & 94& 0.001&96 & 0.03\\    
                & 70 & 1.0 & 100.12 & 77.0 & 84.98 & 96& 0.001&98 & 0.03\\       
         Grid & 10 & 0.0 & 319.00 & 0.0 & 319.00 &  15 & 0.01& 5 & 0.19\\
         10 goals & 30 & 0.0 & 427.90 & 0.0 & 427.90 &  29& 0.01& 11 & 0.19\\    
                  & 50 & 0.0 & 361.14 & 0.0 & 361.14 &  55& 0.01& 19 & 0.19\\    
                  & 70 & 0.0 & 312.50 & 0.0 & 312.50 &  59& 0.01& 21 & 0.19\\       \midrule    
         Buy & 10 & - & - & - & - & 100& 0.0003 &80 & 0.19\\
          2 goals & 30 & - & - & - & - &100 & 0.0003& 80 & 0.19 \\    
          & 50 & - & - & - & - & 100&  0.0003& 80 & 0.19\\    
          & 70 & - & - & - & - & 100&  0.0003 & 80 & 0.19\\   
         Buy & 10 & - & - & - & - &12 & 0.002 & 10& 0.26 \\ 
          10 goals & 30 & - & - & - & - & 34& 0.002 & 9& 0.26\\     
          & 50 & - & - & - & - & 55& 0.002& 9& 0.26 \\     
          & 70 & - & - & - & - & 74& 0.002& 9& 0.26\\     
          \bottomrule
        \end{tabularx}
        \end{minipage}
        \caption{Comparison of LSTM, XGBoost, LGR and LGR+prior information with respect to accuracy (Acc. columns) and
          time to perform goal recognition in several domains under some observability ratios (Observ. column).}
    \label{tab:results}
\end{table*}

\begin{figure*}[hbtp]
    \centering
    \includegraphics[width=0.32\linewidth]{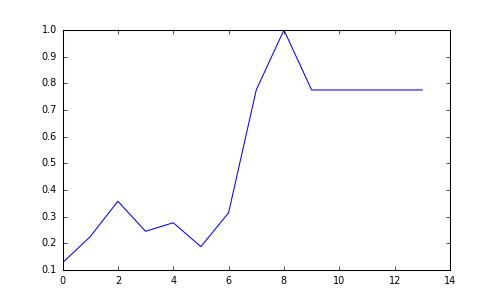}
    \includegraphics[width=0.32\linewidth]{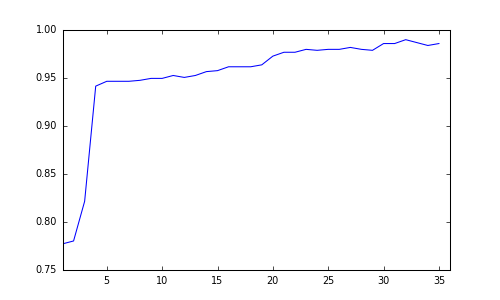}
    \includegraphics[width=0.32\linewidth]{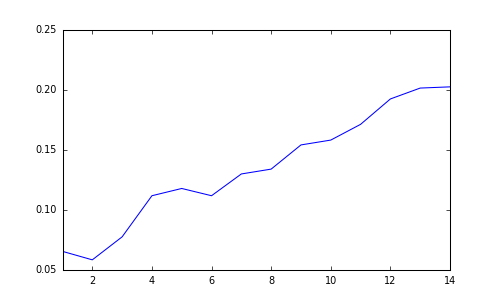}
    \caption{Convergence results for the LSTM model in three cases corresponding to the Block-words domain fixed initial state (left)
    as well as the Grid domain fixed initial state problems (middle) and the Grid domain problem with random initial states with $10$ goals (right). Typically, the performance increases as the model observes more actions  except for one of the cases where we see that towards the end it drops back to $80$ percent. x-axis represent the number of traces used for training, and the y-axis represents the corresponding accuracy.}
    \label{fig:lstm_conv}
\end{figure*}

In both settings of the Grid domain, the LGR method did not perform well since there are no landmarks in this domain. Other methods for goal recognition such as
computing optimal plans that satisfy the observations could be used instead. However, computing optimal plans on partial observations to all possible goals can be prohibitively expensive. Additionally, human behavior may not be optimal, and so assuming rationality in human goal recognition maybe misleading. 
As for the LSTM's performance in this domain, it is worse than XGBoost. We hypothesize that since XGBoost sees the data in sets, rather than sequences (as LSTMs), it works in it's favor; the variety of paths in the Grid domain makes learning sequential patterns more difficult. The lack of clear landmarks hurts the LSTM method too. This issue should be explored in the future.

In the Block-words domain, specifically for the setting in which there were 2 goals, the LSTM settled on predictions using the data distribution, i.e. predicting the most likely goal. XGBoost did better than LSTM. We think this maybe for a similar reason as in the Grid domain; the LSTM could not easily learn any clear action sub-sequences to use as predictive features, especially since the 2 goals were very close. XGBoost, on the other hand, is looking at traces as sets of actions rather than as sequences, so it does not suffer from long common prefix sequences. XGBoost also did better than LGR, possibly because the goals are very close to each other. In Block-words, LGR could not easily find landmarks for very similar goals with fewer actions in the traces. As the number of observations increases, LGR does better than LSTM, but still worse than XGBoost. However, in the setting where there were 5 goals, and the initialization was to random states, the LSTM did worse with fewer observations, but increased its performance with more observations. We hypothesize that the large number of prefix plans arising from random initial conditions made pattern detection difficult with fewer observations. Both XGBoost and LGR still fared better than LSTM with few observations. A key requirement for using XGBoost is that the maximum number of actions must be known and fixed, whereas the LSTMs allow for a variable number of actions. 

The Buy domain was designed to be difficult, especially in the case of 10 goals as the results show. LGR could not solve any problem given that it cannot handle numeric variables. Again, XGBoost performs better than LSTMs, probably again for the same reasons as before. The prefix of the observation traces were very similar with many payroll actions before committing to a specific goal. This kind of domain is not appropriate for planning-based approaches given that the rational of which goal is closer to the plan does not depend on the observed actions.

With respect to the results for the Logistics domain, we noticed a stark difference in performance between the case with two goals and the case with 10 goals. When there were only two goals and the same fixed initial location of the packages, the LSTM and XGBoost methods were able to perform similarly to the LGR method. XGBoost was able to detect and adjust its predictions based on the discerning actions in the tail end of the traces. Since the plans to the two goals had a lot of shared actions in the prefix, XGBoost probably learned to ignore them, and paid attention to the tail end of the traces. This is seen empirically: XGBoost's accuracy increases when the number of actions increases from $50\%$ to $70\%$. LSTM on the other hand seemed to converge to predicting based on the distribution of the goals, and stayed there. 
However, the performance for the Logistics domain setting with 10 goals was quite different. In this setting, the initial state was random, so there was a great diversity in the plan traces, and this made learning patterns very difficult. The actions in Logistics have a partial order that allows for a greater variety of plan traces. This makes finding and utilizing sequence-based patterns harder. So the LGR method performed much better than the LSTM-based approach for Logistics with 10 goals. So, contrary to our expectations, the LSTM learning-based approach did not always guarantee better results than a model-based approach. For some other domains, the order of actions is more strict and so learning-based methods would have an easier time finding patterns to use for goal prediction.

We did try to coax the LSTM method into better performance with fine tuning the hyper-parameters and increasing model capacity. Even with 50 epochs of training, the accuracy was only as high as about $47\%$ and that too only after seeing over half of the entire plan trace. Doubling the size of the hidden dimensional space of the LSTM from 32 to 64 increased the accuracy by less than $1\%$. So, when the data complexity is high, or rather the partial ordering of actions allows for a greater variety of plan traces, using a principled model-based approach seems to do better. The trade-off is that model-based approaches require more computation time during inference as seen in Table~\ref{tab:results}, while learning-based approaches need more computation time during training.

Another issue has to do with model complexity. Model-based methods are either prohibitively expensive for numeric domains or simply do not support them. In such cases, learning-based methods might be the only alternative, or at least provide better performance. This gap in model-based goal recognition is an opportunity for research, and will be needed for financial domains.

\section{Conclusions and Future Work}

Goal recognition on plan traces of actions can be used in finance to better understand customer goals from their actions, and then provide better services to them. Goal recognition can be done through model-based or model-free (learning) methods. In this paper, we have adapted state-of-the-art model-free techniques to work for goal-recognition and compared their performance with that of state-of-the-art model-based approaches. We have performed this comparison using different domains: three standard IPC domains; and one novel domain geared towards mimicking the use of goal recognition for financial tasks. We have analyzed the results and discussed the trade-offs between the two types of approaches. 

The obtained results reinforce the knowledge on the differences between the two main AI paradigms. The two main known differences are: model-based need a model, which requires some knowledge engineering effort, while model-free require a potentially huge amount of examples and training effort. Also, model-based approaches typically assume agents' rationality, while model-free methods automatically adapt to the particularities of observed agents.

One of the new conclusions is that model-based approaches are better when there is a partial order of actions in plans. This feature allows for a great diversity of plan traces, which makes learning action-patterns for goal predictions very hard for learning-based methods, more specifically to sequence-based ones, as LSTMs. As expected, this comes at the cost of computation time during inference. On the other hand, if there is a relation between some actions (not directly associated to goal achievement) and the goals, learning techniques will be able to capture that relation, while model-free will fail to recognize it and present worse performance. 

In future work, we would like to study hybrid approaches that combine model-based and model-free approaches to goal recognition, and how to combine them for different settings. We would also like to investigate the performance of different goal-recognition techniques under noisy observability, goal-recognition under plan obfuscation from an opponent planning-based system~\cite{Kulkarni20}, and plan completion for subsequent goal recognition. We would also like to use attention-based models such as transformers~\cite{vaswani2017attention} for goal recognition, which has hitherto not been done.








\section*{Acknowledgements}

This paper was prepared for information purposes by the Artificial Intelligence Research group of JPMorgan Chase
  \& Co. and its affiliates (``JP Morgan''), and is not a product of the Research Department of JP Morgan.  JP Morgan
  makes no representation and warranty whatsoever and disclaims all liability, for the completeness, accuracy or
  reliability of the information contained herein.  This document is not intended as investment research or investment
  advice, or a recommendation, offer or solicitation for the purchase or sale of any security, financial instrument,
  financial product or service, or to be used in any way for evaluating the merits of participating in any transaction,
  and shall not constitute a solicitation under any jurisdiction or to any person, if such solicitation under such
  jurisdiction or to such person would be unlawful.  \textcopyright 2020 JPMorgan Chase \& Co. All rights reserved

\bibliographystyle{named}

\end{document}